\documentclass{article}
\usepackage{spconf,amsmath,graphicx}
\usepackage{booktabs}
\usepackage{color,cite,balance,url,siunitx}
\usepackage{multicol,multirow}
\usepackage{amssymb}

\usepackage[table]{xcolor}
\usepackage{pifont}
\usepackage{threeparttable}

\newcommand{\cmark}{\ding{51}}
\newcommand{\xmark}{\ding{55}}

\title{An Exploration of Self-Supervised Pretrained Representations for End-to-End Speech Recognition}
\name{
\begin{tabular}{c}
\it Xuankai Chang$^1$, Takashi Maekaku$^{2*}$, Pengcheng Guo$^{3*}$, Jing Shi$^{4*}$, \\
\it Yen-Ju Lu$^{5}$, Aswin Shanmugam Subramanian$^{6}$, Tianzi Wang$^{6}$, Shu-wen Yang$^{7}$, \\
\it Yu Tsao$^{5}$, Hung-yi Lee$^{7}$,  Shinji Watanabe$^{1}$ \\
\end{tabular} \thanks{$^*$Equal contribution.}
}

\address{
    $^1$ Carnegie Mellon University, $^2$ Yahoo Japan Corporation, $^3$ Northwestern Polytechnical University \\
    $^4$ Institute of Automation, Chinese Academy of Sciences, $^5$ Academia Sinica, $^6$ Johns Hopkins University \\
    $^7$ National Taiwan University \\
}
\begin{document}
\ninept
\maketitle
\begin{abstract}
Self-supervised pretraining on speech data has achieved a lot of progress. High-fidelity representation of the speech signal is learned from a lot of untranscribed data and shows promising performance. Recently, there are several works focusing on evaluating the quality of self-supervised pretrained representations on various tasks without domain restriction, e.g. SUPERB. However, such evaluations do not provide a comprehensive comparison among many ASR benchmark corpora.
In this paper, we focus on the general applications of pretrained speech representations, on advanced end-to-end automatic speech recognition (E2E-ASR) models. We select several pretrained speech representations and present the experimental results on various open-source and publicly available corpora for E2E-ASR. Without any modification of the back-end model architectures or training strategy, some of the experiments with pretrained representations, e.g., WSJ, WSJ0-2mix with HuBERT, reach or outperform current state-of-the-art (SOTA) recognition performance. Moreover, we further explore more scenarios for whether the pretraining representations are effective, such as the cross-language or overlapped speech. The scripts, configuratons and the trained models have been released in ESPnet to let the community reproduce our experiments and improve them.
\end{abstract}
\section{Introduction}
\label{sec:intro}

The performance of speech recognition systems have been improved a lot over the last decade. On the one hand, the rapid development of deep neural networks has dramatically pushed the limit of the models~\cite{abdel2012applying,graves2013speech,chan2015listen,vaswani2017attention,dong2018speech,gulati2020conformer}. On the other hand, the increasing computing resources have enabled to train an automatic speech recognition (ASR) system with a large amount of transcribed data~\cite{panayotov2015librispeech,chen2021gigaspeech}, leading to a better performance. It is known that the deep neural networks are data hungry, thus some researchers have been trying to improve the capacity of neural networks by incorporating more and more transcribed data~\cite{chan2021speechstew}. However, using the transcribed data only is not efficient because the untranscribed data is of great portion in all the data available. Motivated by this, researchers proposed to make use of the untranscribed data, known as unsupervised and semi-supervised learning. Recently, it has become a hot topic in speech recognition and can be roughly divided into two approaches. In~\cite{lee2013pseudo,synnaeve2019end,kahn2020self}, a semi-supervised method, called pseudo-labelling, was proposed to use both transcribed and untranscribed data. A seed model is first trained with the transcribed data in a supervised manner, which is then used to generate the pseudo-labels for the untranscribed data. After that, a model can be trained with all the data in a supervised manner.

Previous studies in computer vision (CV) and natural language processing (NLP) have investigated to learn representations from data, showing the advantages in the corresponding downstream tasks~\cite{vinyals2016show,devlin2018bert,radford2018improving}. Similarly, another approach was proposed to pretrain models using a large amount of untranscribed data for learning high quality speech features. In this context, many pretrained speech representation models have been proposed, which are often referred as self-supervised learning representation (SSLR). These SSLRs can be categorized by their training objectives. To be specific, one direct way to learn the speech representations is to predict the future information given the history information. In~\cite{chung2019unsupervised,chung2020vector}, the authors adopted an method similar to the autoregressive language models (LMs) to predict the future acoustic features (e.g. FBANK) conditioned on the past input features, called autoregressive predictive coding (APC). Instead of autoregressive modeling, some researchers proposed to use masking prediction techniques as in BERT-LM~\cite{devlin2018bert} to learn the speech representations, including Mockingjay~\cite{liu2020mockingjay}, TERA~\cite{liu2020tera} and NPC~\cite{liu2020non}. However, it is not necessary to learn the speech representations by reconstructing the acoustic features. In~\cite{oord2018representation,schneider2019wav2vec}, the models were optimized with a contrastive loss to distinguish the positive sample from negative samples in predictions of future. Later in~\cite{baevski2019vq, baevski2020wav2vec}, a BERT Transformer model is concatenated after the encoder trained by the contrastive loss. Recently, a novel model, called HuBERT~\cite{hsu2021hubert}, was proposed to pretrain the representation model by a classification tasks using pseudo-labels motivated by deep cluster models \cite{caron2018deep,xie2016unsupervised}.

All the proposed representations have shown promising results, however, we can hardly draw a conclusion about a suitable representation for various tasks because their experiments focused on a limited number of tasks and were performed independently. Recently, a benchmark, called Speech processing Universal PERformance Benchmark (SUPERB)~\cite{yang2021superb}, was proposed to provide a fair and standard evaluation of various speech representations, with a unified toolkit, S3PRL. SUPERB focuses on the shallow information of each representation. During evaluation, all the representation models are frozen and applied on several downstream tasks each of which uses a quite light-weight downstream model. For example, the ASR task is evaluated using a two-layer RNN-based connectionist temporal classification (CTC) model. Such evaluation provides informative clues to compare the capacity and the concentration of information for each SSLR. Besides, it prepares the easy access to a lot of pretrained SSLR models. Thus, SUPERB is a very strong benchmark for evaluating the SSLRs without any doubt. 

With that being said, it still remains a question that how well these SSLRs can perform in the advanced speech recognition systems. In this paper, we investigate the performance of end-to-end ASR (E2E-ASR) systems using the pretrained SSLRs. To achieve this, we incorporated the SSLRs from S3PRL, the toolkit used in SUPERB, to the ESPnet\cite{watanabe2018espnet}, a widely used E2E speech processing toolkit. Thus, we can easily evaluate the E2E-ASR performance of pretrained SSLRs available in S3PRL using the current state-of-the-art (SOTA) neural network models, such as Transformers~\cite{vaswani2017attention,dong2018speech} and Conformers~\cite{gulati2020conformer}. We can also easily evalute the SSLRs in other downstream tasks, including speech translation (ST) \cite{inaguma2020espnet} and speech enhancement (SE) \cite{li2021espnet}. It is also an interesting question in the air about the generalization ability of these SSLRs, given the fact that most of SSLRs were trained and tested mainly on LibriSpeech~\cite{panayotov2015librispeech,kahn2020libri}. In this project, we explored these pretrained SSLRs on various open-source and publicly available corpora as many as possible, considering different characteristics including read vs. spontaneous speech, single-speaker vs. multi-speaker, noisy/distant-talk environments, and the telephone channel. We show that some of the SSLRs can achieve much better results than the commonly used log-Mel Filterbank (FBANK) feature.

The contributions of this study include:
\begin{itemize}
    \item We implement the use of pretrained SSLRs in advanced E2E-ASR models, based on which we compare the performance of different representations.
    \item The experimental results show that simply replacing the FBANK features with the SSLRs can surpass our current best E2E-ASR system. For some ASR benchmark corpora, our results get competitive results with the SOTA systems, such as WSJ, LibriSpeech and TEDLIUM2.
    \item We explore more scenarios with domain-mismatch from the raw speech data to train the pretraining representations. Some observations show the relationship between pretraining representations and their applicable scenarios.  
    \item We provide reproducible benchmark results, recipes, setups and well-trained models on several publicly available corpora in our open source toolkit ESPnet.
\end{itemize}

\section{End-to-End ASR}
\label{sec:asr}
In this section, we will briefly describe the E2E-ASR models used in this paper, including the CTC and the attention-based encoder-decoder (AED) framework.

\subsection{CTC}
\label{ssec:ctc}
Giving an input speech feature $\mathbf{X} = (\mathbf{x}_1, \dots, \mathbf{x}_T)$, where $T$ means the number of frames, and the corresponding output label sequence $\mathbf{Y} = (y_1, \dots, y_U)$, where $y_u \in \mathcal{V}$ and $\mathcal{V}$ is a vocabulary, CTC~\cite{graves2006connectionist} was proposed to estimate a frame-level input-output alignment $\pi = (\pi_1, \dots , \pi_T)$ by allowing repetitions of labels or emitting a special blank label $\epsilon$, i.e., $\pi_t \in  \mathcal{V} \bigcup \{ \epsilon \}$. The CTC optimizes the model to maximize the probability distribution $P(\mathbf{Y} | \mathbf{X})$ over all possible alignments, which can be formulated as:
\begin{align}\label{eq:ctc_prob}
    P_{\text{CTC}}(\mathbf{Y} | \mathbf{X}) = \sum_{\pi \in \Phi(\mathbf{Y})} P(\pi | \mathbf{X}),
\end{align}
where $\Phi(\mathbf{Y})$ refers all possible alignments of $\mathbf{Y}$. 

\subsection{Joint CTC/Attention-based Encoder-Decoder}
\label{ssec:aed}
AED model directly maps an input speech feature into an label sequence of words or characters without any intermediate representations. It models the distribution of each label by conditioning on both previous estimated labels and the input feature as:
\begin{align}
    P_{\text{Attn}}(\mathbf{Y} | \mathbf{X}) &= \prod_u P(y_u | \mathbf{X}, y_{1:u-1}). \label{eq:aed_prob}
\end{align}

For better performance, we adopted the joint CTC/attention-based encoder-decoder architecture~\cite{kim2017joint} in this work. The loss is defined as the sum of the negative log-likelihoods of CTC and AED:
\begin{align}
    \mathcal{L} = -\left[\lambda \ln P_{\text{CTC}}(\mathbf{Y} | \mathbf{X}) + (1-\lambda) \ln P_{\text{Attn}}(\mathbf{Y} | \mathbf{X})\right],
\end{align}
where $\lambda \in [0,1]$ is a tunable hyper-parameter.

In this work, we majorly use Transformer~\cite{vaswani2017attention} and Conformer~\cite{gulati2020conformer} as the basic block to build our E2E-ASR models.

\subsubsection{Transformer}
\label{sssec:transformer}
Transformer~\cite{vaswani2017attention} was proposed by Vaswani \textit{et al.} and has become the dominant model in various E2E-ASR tasks~\cite{dong2018speech, karita2019comparative}. Both the encoder and decoder are multi-blocked architectures and each block is stacked by a multiheaded self-attention (MHSA) module and a positionwise feed-forward (FFN) module. For the decoder, another source-target attention module is inserted after the MHSA module to joint model the acoustic and linguistic information. 

\subsubsection{Conformer}
\label{sssec:conformer}
In~\cite{gulati2020conformer}, Gulatiet \textit{et al.} proposed a novel architecture with combination of self-attention and convolution in ASR models, which is named Conformer encoder. Specifically, it includes a MHSA module, a convolution (CONV) module, and a pair of FNN modules in the Macaron-Net style. While the MHSA learns the global context, the CONV module efficiently captures the local correlations synchronously. In this work, our model follows the same setups as in~\cite{guo2021recent}, which integrates the Conformer encoder with a normal Transformer decoder.

\section{Speech Representations for ASR}
\label{sec:feature}

Basically, raw speech signal is much less efficient in conveying information than text. Thus for ASR, speech representation extraction is an important module to condense the information of speech signal. Traditionally, many handcrafted features, such as log-Mel Filterbank (FBANK), Mel Frequency Cepstral Coefficients (MFCCs), are used in ASR tasks. There are many SSLRs proposed in previous studies. In this project, we cover the following 8 typical methods to perform the evaluation, including APC, CPC, HuBERT, Mockingjay, NPC, TERA, VQ-APC, Wav2Vec2.0\footnote{We only evaluated Wav2Vec2.0 among the Wav2Vec series because we believe it can represent the best of Wav2Vec techniques.}. 

\begin{table*}[htbp!]
    \centering
    \caption{Information summary of the SSLRs used in this paper, including the data used in pretaining (Data), model architecture (Arch), number of parameters (\#Params) and the stride of the features.}
    \resizebox{1.0\linewidth}{!}{
    \begin{tabular}{c|cccc|cccc}
        \toprule
        \multirow{2}{*}{SSLR} & \multicolumn{4}{c|}{Objective} & \multirow{2}{*}{Data} & \multirow{2}{*}{Arch} & \multirow{2}{*}{\#Params} & \multirow{2}{*}{Stride} \\
         & autoregressive & masking & contrastive & pseudo-labelling &  &  &  & \\
        \hline\hline
        APC        & \cmark &  &  &  & LibriSpeech 960hr & 3-GRU & 4.1M & 10ms \\
        CPC        &  &  & \cmark &  & LibriLight 60K & 5-Conv, 1-LSTM & 1.8M & 10ms \\
        HuBERT     &  &  &  & \cmark & LibriLight 60K & 7-Conv 24-Trans & 316.2M & 20ms \\
        Mockingjay &  & \cmark &  &  & LibriSpeech 960hr & 3-Trans & 21.3M & 10ms \\
        NPC        &  & \cmark &  &  & LibriSpeech 960hr & 4-Conv, 4-Masked Conv & 19.4M & 10ms \\
        TERA       &  & \cmark &  &  & LibriSpeech 960hr & 3-Trans & 23.3M & 10ms \\
        VQ-APC     & \cmark &  &  &  & LibriSpeech 960hr & 3-GRU & 4.6M & 10ms \\
        Wav2Vec2.0   &  &  & \cmark &  & LibriSpeech 960hr & 7-Conv 24-Trans & 317.4M & 20ms \\
        \bottomrule
    \end{tabular}
    }
    \label{tab:sslrs}
\end{table*}


\subsection{Autoregressive-prediction based representations}
\label{ssec:apcs}
One type of representation learning method is to learn to predict the future acoustic features given the past.

\noindent \textbf{APC}. 
Autoregressive Predictive Coding (APC) was proposed in \cite{chung2019unsupervised}. The idea comes from autoregressive LMs for text, which is typically a probability distribution over token sequences. APC uses an RNN for modeling the temporal information of raw speech signals to predict the features of the frame $K$ steps ahead. The model is optimized by minimizing the L1 loss between the input speech signal and the predicted sequence.

\noindent \textbf{VQ-APC}.
VQ-APC \cite{chung2020vector} is a variant of APC that incorporates vector quantization (VQ) layers. A VQ layer in the middle of the APC network is used to control the amount of information from the previous layer. Therefore, the model is forced to learn better representations to predict future frames. 

\subsection{Masking-prediction based representations}
\label{ssec:mask-codings}

Speech signal is context dependent. Using both the past and future information when learning the speech representation can be useful. Some researchers proposed to generate the current frame conditioned on both the past and future information.

\noindent\textbf{Mockingjay}. Similar to BERT\cite{devlin2018bert}, the model is trained by recover the masked input features based on its left and right context frames. During training, a certain amount of input frames are selected to be dynamically masked by zero-masking, random value filling or leaving unchanged. The model is optimized by minimizing the reconstruction loss between the original feature sequence and the prediction using L1 loss. To avoid the model learning the local smoothness, the frames being masked are chosen as a consecutive frame subsequence. The MFCC feature is used as the acoustic features. A Transformer with multi-headed self-attention is used as the model. On top of it, a two-layer feed-forward network was used to predict the original feature.

\noindent\textbf{TERA}. Transformer encoder representation from alteration (TERA) \cite{liu2020tera} is an extension of Mockingjay. Similar to Mockingjay, the model takes the manipulated acoustic features as input and minimizes the difference between the ground-truth and the prediction at masked portion of the input. In addition to the masking process along time axis used in Mockingjay, TERA employs two more alterations, including the masking process in frequency axis and the magnitude alterations by adding random noise.

\noindent\textbf{NPC}. Nonautoregressive predictive coding (NPC) \cite{liu2020non} shares the similar principle with Mockingjay and TERA. For NPC, it uses the convolutional neural networks. The model generates the acoustic feature at each time step $t$ conditioned on both the future and past information within a certain range. To avoid the local smoothness problem, the nearest frames within $m$ steps are masked out. Thus, the input of the model is in the range $[t-r, t+r] - [t-m, t+m]$.

\subsection{Contrastive learning based representations}
\label{ssec:cpcs}
\textbf{CPC}.
Instead of using a conditional generative model to reconstruct the original input signal, the contrastive predictive coding (CPC) \cite{oord2018representation} model learns the representation via maximizing the mutual information between the current context and future embeddings by minimizing the noise-contrastive estimation-based (NCE) loss\cite{gutmann2010noise}. We use an updated version of CPC model, called modified-CPC \cite{riviere2020unsupervised}. The model contains a 5-layer convolutional neural network (CNN) encoder to generate latent representation at lower temporal resolution and 1-layer LSTM to summarize the latent representation as a context representation. 

\noindent\textbf{Wav2Vec2.0}.
Wav2Vec2.0 \cite{baevski2020wav2vec} is also a contrastive learning-based approach. It is composed of a CNN-based encoder network, a VQ module, and a Transformer-based context representation network. The masking idea is also applied in the model, the input to the Transformer being randomly masked. During the pretaining, Wav2Vec2.0 is optimized with a contrastive loss function. In addition, a regularization term is added to the loss function to increase the diversity of the codebook in the VQ module.

\subsection{HuBERT}
\label{ssec:hubert}
\textbf{HuBERT} 
Hidden-Unit BERT(HuBERT) \cite{hsu2021hubert} is another novel self-supervised speech representation learning approach. During pretaining, iterative pseudo-labelling method is adopted. First, offline clustering method (e.g. k-Means) is applied on the acoustic features of untranscribed speech, such as MFCC. In this process, the discrete labels is generated for the speech data. There are two major modules in the pretained HuBERT model including a CNN-based encoder and a BERT encoder. The HuBERT model is trained to predict the distribution of discrete units. Similar to BERT-LM, the input feature to the BERT encoder is partially masked. The cross-entropy loss is computed on both the masked and unmasked frames. After training, the model can be used to generate the new pseudo-labels with higher quality. Such refinement can be carried out for many times.

\subsection{Application of SSLRs}
\label{ssec:app}
\begin{figure}
    \centering
    \includegraphics[width=\linewidth]{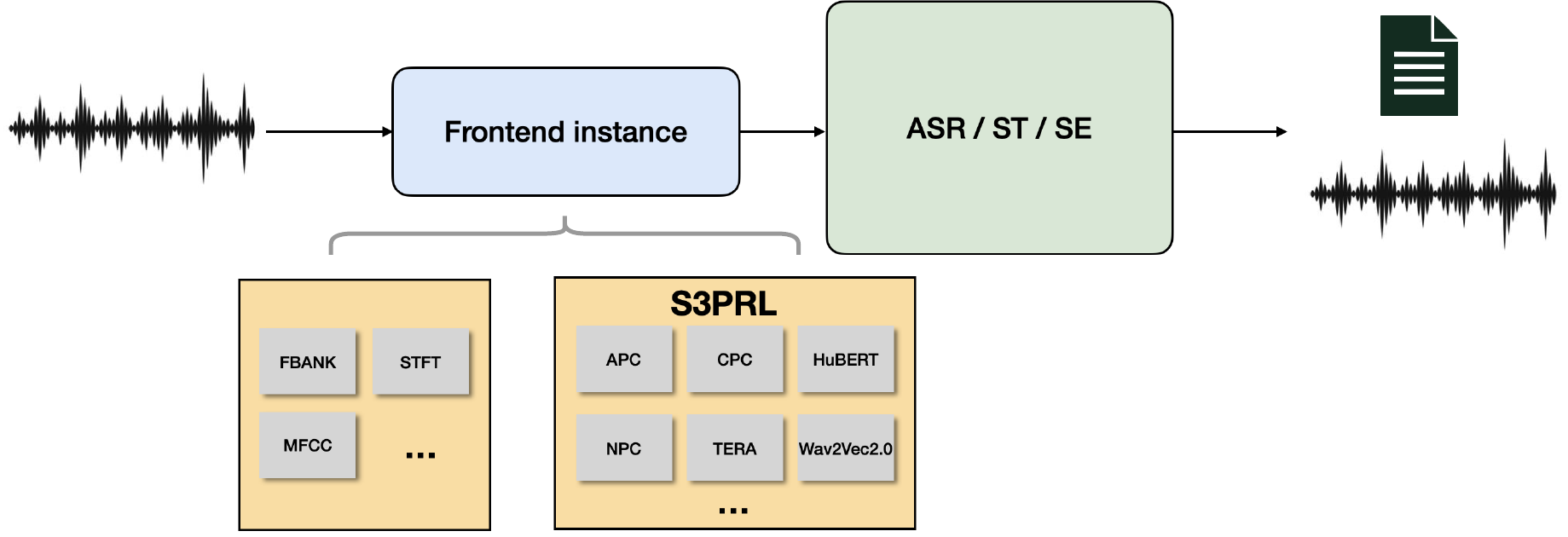}
    \caption{End-to-End speech processing with various speech representations. The framework can be used in speech recognition (ASR), speech translation (ST), speech enhancement (SE), etc.}
    \label{fig:model}
\end{figure}
All the SSLR models we use are pretained and can be accessed via S3PRL. For HuBERT and CPC, we used the pretained model with 60,000h Libri-Light\cite{kahn2020libri}. For the rest of the SSLRs, we use the models pretained with LibriSpeech 960h \cite{panayotov2015librispeech}. The detailed information of SSLRs is described in Table~\ref{tab:sslrs}. We follow the policies in SUPERB\cite{yang2021superb} to use the weighted-sum of multiple hidden states as the input to E2E-ASR. A weight parameter is trained to summarize all the hidden states for each SSLR model. The overall E2E-ASR is shown in Fig.~\ref{fig:model}. The input speech goes through a frontend to extract feature. Various of SSLRs can be extracted instead of conventional features, such as FBANK. Next, the feature sequence is fed into the ASR network to predict the hypothesis.

\section{Experiments}
\label{sec:illust}

\subsection{Setups}
\label{ssec:setup}
To evaluate the performance of various self-supervised learning representations, we conduct experiments on several open-source and publicly available corpora. Although the pretained representation models were learned on the LibriSpeech 960 or LibriLight data, we tested their performance on other datasets to verify their generalization ability. We conducted experiments on 7 ASR corpora \footnote{We did our best to work on as many datasets as possible. Due to the time limit, we couldn't finish all the experiments before submission unfortunately. We are still working on some other corpora including but not limited to REVERB\cite{kinoshita2013reverb}, AMI\cite{mccowan2005ami}, VoxForge, HKUST\cite{liu2006hkust}, TEDLIUM3\cite{hernandez2018ted}.}. We try to cover various aspects in the ASR task, including read vs. spontaneous speech, noisy/distant-talk environment and the telephone channel. 

Basically, we follow the recipes in the latest ESPnet\cite{watanabe2018espnet} to conduct the experiments. The default FBANK feature extractor is replaced with the pretained self-supervised learning representation models. The downsampling layers are used to make all the SSLRs have a 40ms-stride in the encoder. For most of the corpora, we use the Conformer encoder and Transformer decoder as our network architecture, which showed promising results \cite{guo2021recent}. The hyper-parameters of the ASR models are shown in Table~\ref{tab:setups}.

\begin{table}[htbp!]
    \centering
    \caption{Hyperparameters of the ASR models for each dataset, including data augmentations, encoder type and kernel size (kernel) of Conformer encoder, number of encoder layers ($n_\text{enc}$), number of decoder layers ($n_\text{dec}$), number of attention heads ($H$), feed-forward dimension ($d^\text{ff}$), attention dimension ($d^\text{att})$.}
    \resizebox{1.0\linewidth}{!}{
    \begin{tabular}{c|ccccccccc} 
    \toprule
    \multirow{2}{*}{Dataset} & \multicolumn{2}{c}{DataAugment} & \multirow{2}{*}{EncoderType} & \multirow{2}{*}{kernel} & \multirow{2}{*}{$n_\text{enc}$} & \multirow{2}{*}{$n_\text{dec}$} & \multirow{2}{*}{$H$} & \multirow{2}{*}{$d^\text{ff}$} & \multirow{2}{*}{$d^\text{att}$}  \\ 
     & SpeedPerturb & SpecAug & & & & & & \\
    \hline\hline
    AISHELL & \cmark & \cmark & Conformer & 15 & 12 & 6 & 4 & 2048 & 256\\
    CHiME4 & \xmark & \cmark & Conformer & 15 & 12 & 6 & 4 & 2048 & 256 \\
    LibriSpeech & \cmark & \cmark & Conformer & 31 & 12 & 6 & 8 & 2048 & 512 \\
    Switchboard &\cmark & \cmark & Conformer & 31 & 12 & 6 & 4 & 2048 & 256 \\
    TEDLIUM2 & \cmark & \cmark & Conformer & 15 & 12 & 6 & 4 & 2048 & 256 \\
    WSJ & \xmark & \cmark & Conformer & 15 & 12 & 6 & 4 & 2048 & 256 \\
    WSJ0-2mix & \xmark & \xmark & Transformer & - & 12 & 6 & 4 & 2048 & 256 \\
    \bottomrule
    \end{tabular}
    }
    \label{tab:setups}
\end{table}

\subsection{Performance on E2E-ASR tasks}
\label{ssec:general-asr}
In this part, we present the E2E-ASR results of various SSLRs using joint CTC/attention-based encoder-decoder system. The results on different corpora are shown in Table.~\ref{tab:variousASRtasks}.

\begin{table*}[htbp!]
    \centering
    \caption{Performance (WERs / CERs) of joint CTC/attention-based encoder-decoder on various open source ASR corpora.}
    \begin{threeparttable}
    \resizebox{1.0\linewidth}{!}{
        \begin{tabular}{ccc|c|c|cccccccc} 
        \toprule
        \multirow{2}{*}{Dataset} & \multirow{2}{*}{Vocab} & \multirow{2}{*}{Metric} & \multirow{2}{*}{Evaluation Sets} & \multirow{2}{*}{FBANK} & APC & CPC & HuBERT & Mockingjay & NPC & TERA & VQ-APC & Wav2Vec2.0  \\
        & & & & & (960h) & (60kh) & (60kh) & (960h) & (960h) & (960h) & (960h) & (960h) \\
        \hline\hline
        AISHELL & Char & CER & dev / test & \textbf{4.4} / \textbf{4.7}  & 6.1 / 6.5 & 4.9 / 5.3 & \textbf{4.4} / \textbf{4.7} & 5.0 / 5.4 & 5.0 / 5.5 & 4.7 / 5.1 & 5.0 / 5.5 & 4.6 / 5.0 \\
        \arrayrulecolor{black!30}\midrule
        \multirow{2}{*}{CHiME4} & \multirow{2}{*}{Char} & \multirow{2}{*}{WER} & \{dt05/et05\}-1ch &  13.6 / 23.2  &  16.8 / 27.4  &  16.2 / 25.8 &  \textbf{11.6} / \textbf{22.8}  & - &  16.8 / 28.0 & 16.8 / 28.0  & 17.1 / 27.4 & 13.5 / 26.1\\
        & & & \{dt05/et05\}-5ch &  9.4 / 15.8  &  11.1 / 18.3  & 10.5 / 17.1 &  \textbf{5.0} / \textbf{10.2}  & - &  11.0 / 18.8 & 10.9 / 18.1  & 11.0 / 18.0 & 6.3 / 12.5\\
        \arrayrulecolor{black!30}\midrule
        \multirow{2}{*}{LibriSpeech} & \multirow{2}{*}{BPE} & \multirow{2}{*}{WER} & \{dev / test\}-clean & 1.7 / 1.9 & 2.4 / 2.6 & 2.2 / 2.4 & \textbf{1.5 / 1.6} & 2.3 / 2.4 & 2.5 / 2.6 & 2.4 / 2.5 & 2.2 / 2.6 & 1.7 / 1.9 \\
        & & & \{dev / test\}-other & 4.2 / 4.2 & 7.3 / 7.5 & 6.2 / 6.4 & \textbf{3.1 / 3.2} & 6.4 / 6.9 & 7.3 / 7.5 & 6.7 / 7.0 & 7.0 / 7.3 & 4.9 / 4.7 \\
        \arrayrulecolor{black!30}\midrule
        Switchboard & BPE & WER & eval2000(callhm/swbd) & 15.6 / 8.4 & 17.7 / 8.9 & 15.7 / 8.4 & 18.1 / \textbf{7.3} & 16.9 / 9.9  & 17.6 / 9.1 & 17.2 / 8.9 & 17.4 / 8.6 & \textbf{14.9} / 7.9 \\
        \arrayrulecolor{black!30}\midrule
        TEDLIUM2 & BPE & WER & dev / test & 9.5 / 8.9 & 9.1 / 8.5 & 9.0 / 8.7 & \textbf{6.4 / 6.2} & 9.2 / 8.6 & 9.3 / 8.5  & 8.9 / 8.4 & 9.7 / 9.0 & 7.6 / 7.4 \\
        \arrayrulecolor{black!30}\midrule
        WSJ & Char & WER & dev93/eval92 & 6.6 / 4.4 & 7.2 / 4.5 & 7.1 / 4.7 & \textbf{3.0 / 1.5} & 6.8 / 4.6 & 7.3 / 4.8 & 6.3 / 4.4 & 7.5 / 4.6 & 3.7 / 2.1 \\
        \arrayrulecolor{black!30}\midrule
        WSJ0-2mix & Char & WER & dev / test &  17.7\tnote{$\dagger$} & 16.5  & 14.9 & \textbf{12.1}&  15.9 & 16.5 & 15.1 & 17.1 & 13.2 \\
        \arrayrulecolor{black}\bottomrule
        \end{tabular}
    }
    \end{threeparttable}
    \begin{tablenotes}
        \footnotesize
        \item $\dagger$: The FBANK result is obtained 
    \end{tablenotes}
    \label{tab:variousASRtasks}
\end{table*}

\subsubsection{Read English Speech}
LibriSpeech~\cite{panayotov2015librispeech} is a corpus of English speech extracted from audiobooks. Here we used all 960 hours of data as the training set. We can see that HuBERT, using 60,000h data, outperformed the baseline\footnote{For APC, NPC, and TERA, the number of gradient accumulation was set incorrectly, thus these results could be worse than they actually are. We are re-training the corresponding models.}. In particular, HuBERT achieved a relative improvement of as much as 23.8\% and 26.2\% for \textit{dev-other} and \textit{test-other} sets, respectively. On the other hand, no improvement was observed for other models pre-trained using LibriSpeech 960h. In this case, we used the same data for representation learning and ASR task without fine-tuning. Therefore, it is inferred that there was no room for growth in improvement compared to other corpora setups.

WSJ is a reading speech corpus drawn from the Wall Street Journal news text and the total amount of the training set is about 80 hours. From the Table~\ref{tab:variousASRtasks}, we can find that most of SSLRs show similar results compared with the FBANK feature except HuBERT and Wav2Vec2.0. When simply using the speech representations learned from HuBERT (60kh) and Wav2Vec2.0 (960h), we obtain superior WERs of 3.0\%/1.5\% and 3.7\%/2.1\% on the dev93 and eval92 sets, respectively, reaching the SOTA results (1.3\% on eval92 set) presented in~\cite{chan2021speechstew}, which is trained over 5000 hours English speech.

\subsubsection{Spontaneous English Speech}
We further evaluate the performance with TEDLIUM2\cite{rousseau2014enhancing} dataset, a Semi-Spontaneous English Speech corpus.  Using the HuBERT and Wav2Vec2.0 brings a significant performance gain compared with the FBANK. HuBERT relatively improves the word error rates (WERs) on dev and test sets by $33\%$ and $30\%$, respectively, while Wav2Vec2.0 achieves $20\%$ and $17\%$. For other representations, they achieve better performance than FBANK. Among them, TERA is slightly better than the rest. 

\subsubsection{Noisy English Speech}
CHiME4 \cite{chime4} is a noisy multichannel ASR corpus. This corpus is challenging for self-supervised models because the representations are designed for encoding clean speech signals. We evaluate with both single-mic and multi-mic sets. The 5ch set is enhanced with delay and sum beamforming \cite{anguera2007acoustic}. HuBERT and Wav2Vec2.0 gives significant performance improvement over FBANK features on the multi-mic (5ch) scenario but the performance is similar to FBANK features on the single-mic (1ch) scenario. The other representations degrade the performance compared to FBANK in both scenarios. With Mockingjay, we noticed very serious overfitting issues and it seems inappropriate for this dataset.

\subsubsection{Telephone Channel English Speech}
Switchboard corpus consists of approximately 260 hours of telephone conversation speech and is collected at 8 KHz. Here, we upsample it to 16 KHz to make it suitable for the pretrained models. Since all SSLR models are trained on the reading English speech, there is no doubt that the data mismatches, like speech style, channel distortion, etc., will evidently hinder the performance. From the table, we can find that while most of the SSLR models are susceptible to performance degradation, both HuBERT and Wav2Vec2.0 are robust to such domain mismatches, obtaining a slight improvement compared with FBANK feature.

\subsubsection{Non-English Speech}
Another question remaining is whether these SSLR models are also suitable for cross-lingual scenarios, such Mandarin ASR task Thus, we also conduct experiments on the open-source Mandarin AISHELL corpus~\cite{bu2017aishell}. The AISHELL corpus contains about 178 hours Mandarin speech recorded in a clean environment. Due to the limitation of time and computation resource, we are only able to conduct experiments on a little SSLR models. From the character error rates (CERs) results, we can see that APC and the English language pretrained models are not able to generalize to other languages. Although the HuBERT achieves same results as FBANK feature, other models, like APC and Wav2Vec2.0 shows a severely performance degradation, yielding up to 30\% relative increase of the CERs results.

\subsubsection{Multi-speaker overlapped Speech}
In the previous sections, results on several single-speaker speech corpora show the different performance of the pretaining models. To further explore whether the models, e.g., HuBERT and Wav2Vec2.0, that work well on a single-speaker dataset can be applied in the case of multi-speaker overlapped speech, in this section, we conduct the experiments on the WSJ0-2mix dataset \cite{hershey2016deep}, which is the de-facto benchmark dataset for speech separation or multi-speaker speech recognition systems. In WSJ0-2mix, the 30 hours training set and 10 hours validation set contain two-speaker overlapped mixtures. The 5 hours test set was similarly generated using utterances from another 18 speakers from the WSJ0 validation set and evaluation set which has not overlapped speaker with the training set.

As shown in the last line of Table~\ref{tab:variousASRtasks}, although with the identical architecture of transformer-based model, the results with different pretained representation show a wide range of differences in the term of WER. As shown in the last line of Table~\ref{tab:variousASRtasks}, although with the identical architecture of transformer-based model, the results with different pretained representation show a wide range of differences in the term of WER. To be specific, similar to the previous observation on the single-speaker dataset, the HuBERT and Wav2Vec2.0 also show obvious advantages with the WERs of 12.1\% and 13.5\% respectively. It is worth mentioning that the raw speech data to training the pretained representation are all from single-speaker speech, which is quite different from the overlapped multi-speaker speech. We infer that because of this mismatch between the data distributions, although the back-end models are fully trained with the representation of CPC or Mockingjay, results are not satisfactory.

\begin{figure}[htbp!]
    \centering
    \includegraphics[width=\linewidth]{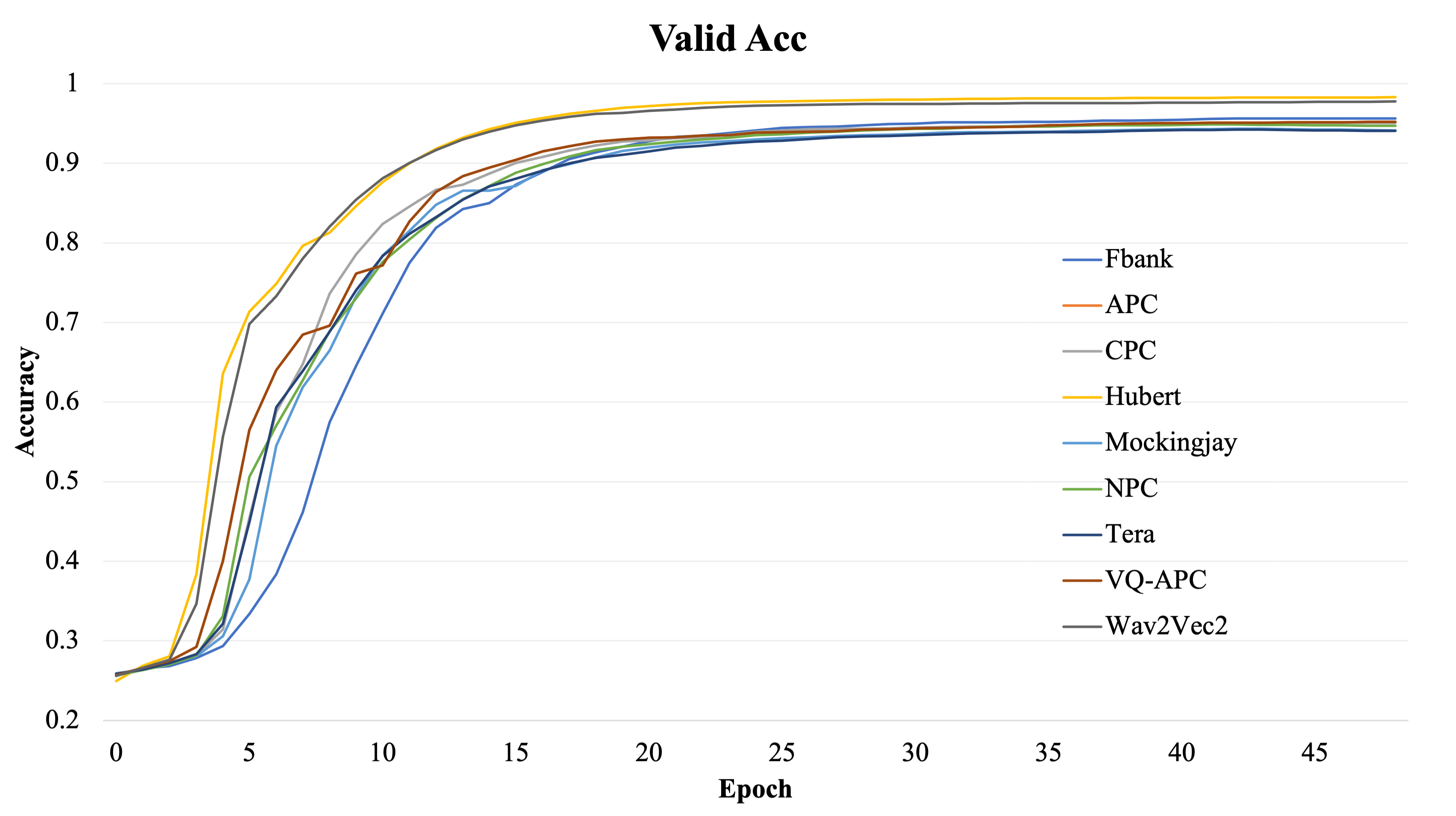}
    \caption{Validation accuracies of different self-supervised learning representations on WSJ ASR dataset.}
    \label{fig:valid_acc}
\end{figure}

\subsection{Performance of E2E Non-autoregressive ASR}
\label{ssec: ctc-aed}
The pretained representations seem to be effective in E2E-ASR from the experiments above, especially the HuBERT and Wav2Vec2.0 representations. It is based on the joint CTC/attention-based encoder-decoder (AED) network. Recently, non-autoregressive (NAR) ASR has become popular. We use the simplest NAR model, namely the CTC-based ASR, to evaluate the performance of every SSLR. The results are shown in Table~\ref{tab:ctcandaed}. When we decode without language model, namely the greedy search in CTC, the HuBERT and Wav2Vec2.0 still outperform all the other representations. The performance of Mockingjay is slightly worse than APC and CPC, different from the joint CTC/AED case. If we add the language model and using beamsearch, the results become much better. For the CTC models, it is surprising that the WERs of HuBERT without LM is even better than those of FBANK with LM.

\begin{table}[htbp!]
    \centering
    \caption{WERs of dev/test sets on WSJ ASR corpora. Comparison between CTC and joint CTC/attention-based encoder-decoder.}
    \resizebox{1.0\linewidth}{!}{
    \begin{tabular}{c|ccc} 
    \toprule
    \multirow{2}{*}{Frontend} & \multicolumn{2}{c}{CTC} & \multirow{2}{*}{Joint CTC \& AED}  \\
     & w/o LM & w/ LM & \\
    \hline\hline
    FBANK & 17.4 / 13.6 & 8.5 / 5.9 & 6.6 / 4.4 \\
    APC & 18.5 / 15.0 & 9.3 / 6.5 & 7.2 / 4.5 \\
    CPC & 18.9 / 14.7 & 9.6 / 6.7 & 7.1 / 4.7 \\
    HuBERT & 8.1 / 5.6 & 3.3 / 2.1 & 3.0 / 1.5 \\
    Mockingjay & 19.2 / 15.6 & 9.7 / 7.0 & 6.8 / 4.6 \\
    NPC & 18.5 / 13.7 & 9.4 / 6.5 & 7.3 / 4.8 \\
    TERA & 18.2 / 14.8 & 8.7 / 6.5 & 6.3 / 4.4 \\
    VQ-APC & 19.4 / 14.7 & 10.0 / 6.5 & 7.5 / 4.6 \\
    Wav2Vec2.0 & 9.4 / 7.2 & 4.3 / 2.5 & 3.7 / 2.1\\
    \bottomrule
    \end{tabular}
    }
    \label{tab:ctcandaed}
\end{table}

\subsection{Weighted-sum vs. Last layer vs. Finetuned Last layer}
\label{ssec: weighted-sum}
\begin{figure}
    \centering
    \includegraphics[width=\linewidth]{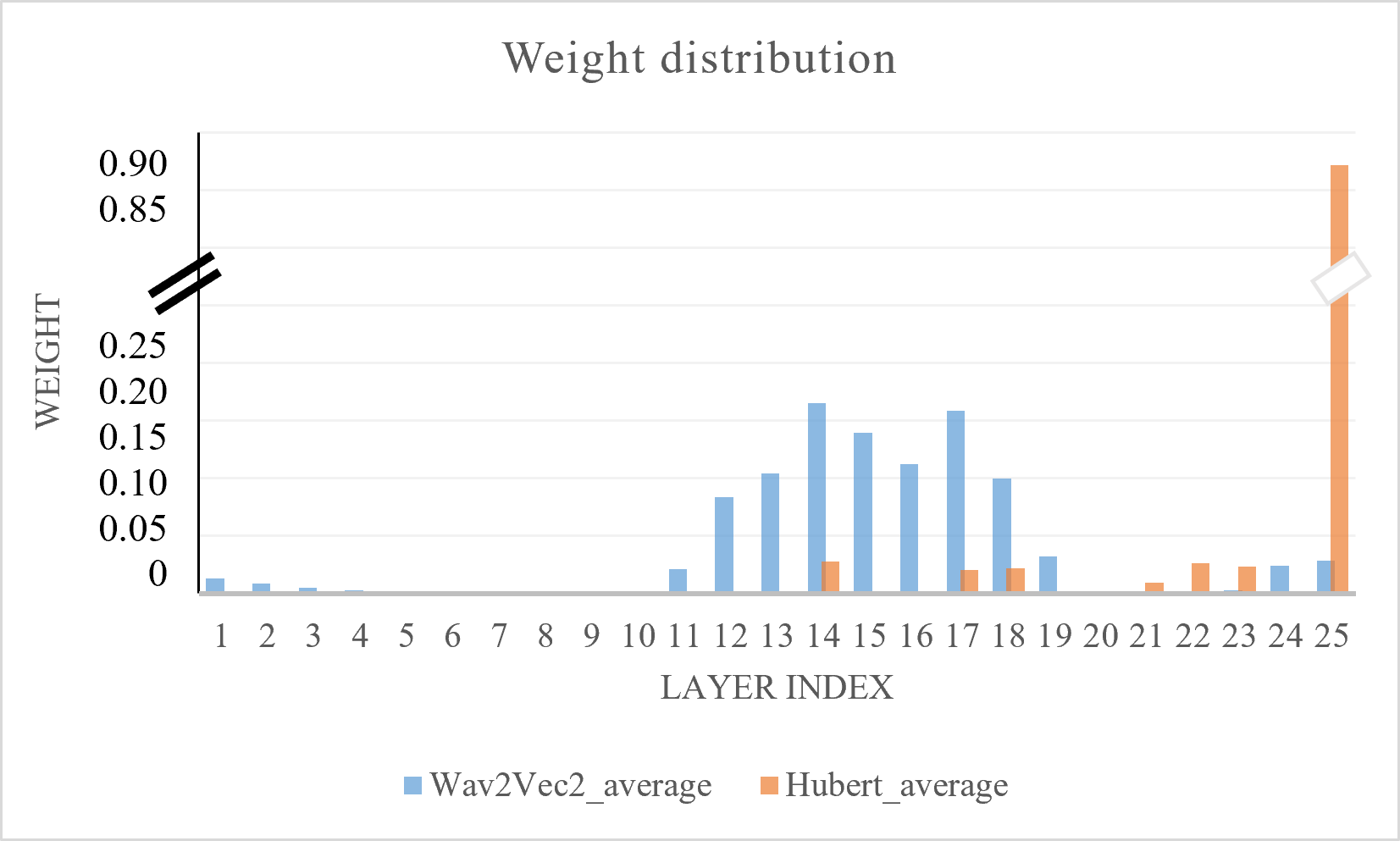}
    \caption{The weights of the feature summarization after training on HuBERT and Wav2Vec2.0 representations.}
    \label{fig:weights}
\end{figure}

In this part, we explore different ways to use the pretained representations. Since the HuBERT and Wav2Vec2.0 achieve much better results than other representations, we mainly evaluate these two models. We test three methods on the WSJ dataset for fast development and the results are presented in Table~\ref{tab:weighted-sum}. To reduce the impact of language models, we show the results of both with LM and without LM. First, we can see that  decoding results of both of the HuBERT and Wav2Vec2.0 without LM are better than FBANK baseline with LM in Table\ref{tab:variousASRtasks}. For Wav2Vec2.0 without LM, the weighted-sum feature performs better than the last-layer cases. This indicates that the hidden states in the middle of the Transformer encoder within Wav2Vec2.0 is also helpful. We further look at the weights of the feature summarization after training, shown in Fig.~\ref{fig:weights}. It can be seen that HuBERT concentrates more on the last-layer. This phenomenon still remains discussion.

\begin{table}[htbp!]
    \centering
    \caption{WERs of dev/test sets on WSJ ASR corpus in w/ and w/o LM conditions. Using weighted-sum, last-layer and finetuned last-layer outputs as ASR input feature.}
    \resizebox{1.0\linewidth}{!}{
    \begin{tabular}{c|cccc} 
    \toprule
    Frontend & LM & Weighted-Sum & Last-layer & Finetuned Last-layer  \\ 
    \hline\hline
    \multirow{2}{*}{HuBERT} & \cmark & 3.0 / 1.5 & 2.6 / 1.6 & 2.8 / 1.6 \\
     & \xmark & 4.9 / 3.4 & 4.8 / 3.4 & 4.6 / 3.5 \\
    \multirow{2}{*}{Wav2Vec2.0} & \cmark  & 3.7 / 2.1 & 3.7 / 2.1 & 3.7 / 2.0 \\
    & \xmark & 6.1 / 3.9 & 6.1 / 4.5 & 6.1 / 4.2 \\
    \bottomrule
    \end{tabular}
    }
    \label{tab:weighted-sum}
\end{table}

\subsection{Discussions}
\label{ssec:discussion}
According to the experiments, we find that the self-supervised learning representations can improve the performance in some cases, especially using the HuBERT and Wav2Vec2.0 models. In Fig.~\ref{fig:valid_acc}, we show the validation accuracies of different SSLRs on the WSJ dataset. All the models use the same optimization parameters. We observed that the learning curve behaviour with the pretained representations are obviously better than FBANK at the first few epochs, which shows the fast convergence properties of all SSLRs.
The accuracies of HuBERT and Wav2Vec2.0 stay the leading position with large margins throughout the training process, which indicates that we can easily judge whether the pretained representations is correctly working or not without waiting for the entire training epochs.

We summarize the other training tips we have observed in our experiment:
\begin{itemize}
    \item If the number of GPUs is insufficient, the accumulating gradient strategy\cite{ott2018scaling} can be employed to emulate a large mini-batch.
    \item When a model suffers from over-fitting, dropout of positional encoding and attentions in Transformer and Conformer block can be enabled.
    \item Strong self-supervised representation models, including HuBERT and Wav2Vec2.0, are robust to optimization parameters, such as the learning rate, batch size (or accumulating gradient), etc. Thus the hyperparameters used in HuBERT and Wav2Vec2.0 may not always fit in others. 
    \item Because HuBERT and Wav2Vec2.0 representations are robust, their training can be used as an upper bound for other representations to monitor the  training trend.
    \item It takes a lot of time to compute the global normalization statistics on CPU. One option is to do it on GPU. For simplicity, we just use utterance-level normalization rather than global one.
\end{itemize}

\section{Conclusion}
\label{sec:conclusion}

In this paper, we explore the application of different pretained self-supervised learning representations in E2E-ASR with the joint CTC/attention-based encoder-decoder architecture. We conduct the experiments on several publicly available corpora. The results show that HuBERT and Wav2Vec2.0, which have the largest number of parameters, can dramatically improve the performance against the commonly used log-Mel Filterbank feature. To achieve better performance, we can directly re-use the existing representation models. We also find that the pretained representations can generalize to other corpora, not restricted to a specific dataset. We plan to open-source all of our configurations and scripts in the ESPnet project, to help the community easily access and improve the performance. In the future, we will extend this project to more corpora and the speech processing tasks beyond ASR.

\section{Acknowledgement}
This work used the Extreme Science and Engineering Discovery Environment (XSEDE) \cite{towns2014xsede}, which is supported by National Science Foundation grant number ACI-1548562. Specifically, it used the Bridges system \cite{nystrom2015bridges}, which is supported by NSF award number ACI-1445606, at the Pittsburgh Supercomputing Center (PSC).

\bibliographystyle{IEEEbib}
\bibliography{strings,refs}

\end{document}